\documentclass[runningheads]{llncs}

\usepackage{eccv}

\usepackage{eccvabbrv}

\usepackage{graphicx}
\usepackage{wrapfig}
\usepackage{color}
\usepackage{booktabs}

\usepackage[accsupp]{axessibility}  % Improves PDF readability for those with disabilities.

\newcommand{\green}[1]{{\color{Green}#1}}
\newcommand{\blue}[1]{{\color{RoyalBlue}#1}}
\newcommand{\orange}[1]{{\color{BurntOrange}#1}}

\usepackage{hyperref}

\usepackage{orcidlink}

\begin{document}

\title{GRLoc: Geometric Representation Regression for Visual Localization} 


\author{
Changyang Li \quad Xuejian Ma \quad Lixiang Liu \quad Zhan Li \quad Qingan Yan \quad Yi Xu}


\institute{Goertek Alpha Labs \\
\email{\{first.last\}@goertekusa.com}
}

\maketitle

\begin{abstract}
  Absolute Pose Regression (APR) has emerged as a compelling paradigm for visual localization. However, APR models typically operate as black boxes, directly regressing a 6-DoF pose from a query image, which can lead to memorizing training views rather than understanding 3D scene geometry. In this work, we propose a geometrically-grounded alternative. Inspired by novel view synthesis, which renders images from intermediate geometric representations, we reformulate APR as its inverse that regresses the underlying 3D representations directly from the image, and we name this paradigm Geometric Representation Regression (GRR). Our model explicitly predicts two disentangled geometric representations in the world coordinate system: (1) a raymap's directions to estimate camera rotation, and (2) a corresponding pointmap to estimate camera translation. The final camera pose is then recovered from these geometric components using a differentiable deterministic solver. This disentangled approach, which separates the learned visual-to-geometry mapping from the final pose calculation, introduces a strong geometric prior into the network. We find that the explicit decoupling of rotation and translation predictions measurably boosts performance. We demonstrate state-of-the-art performance on 7-Scenes and Cambridge Landmarks datasets, validating that modeling the inverse rendering process is a more robust path toward generalizable absolute pose estimation.

\end{abstract}

\section{Introduction}
\label{sec:intro}

\begin{figure}[t]
    \centering
    \includegraphics[width=1.0\textwidth]{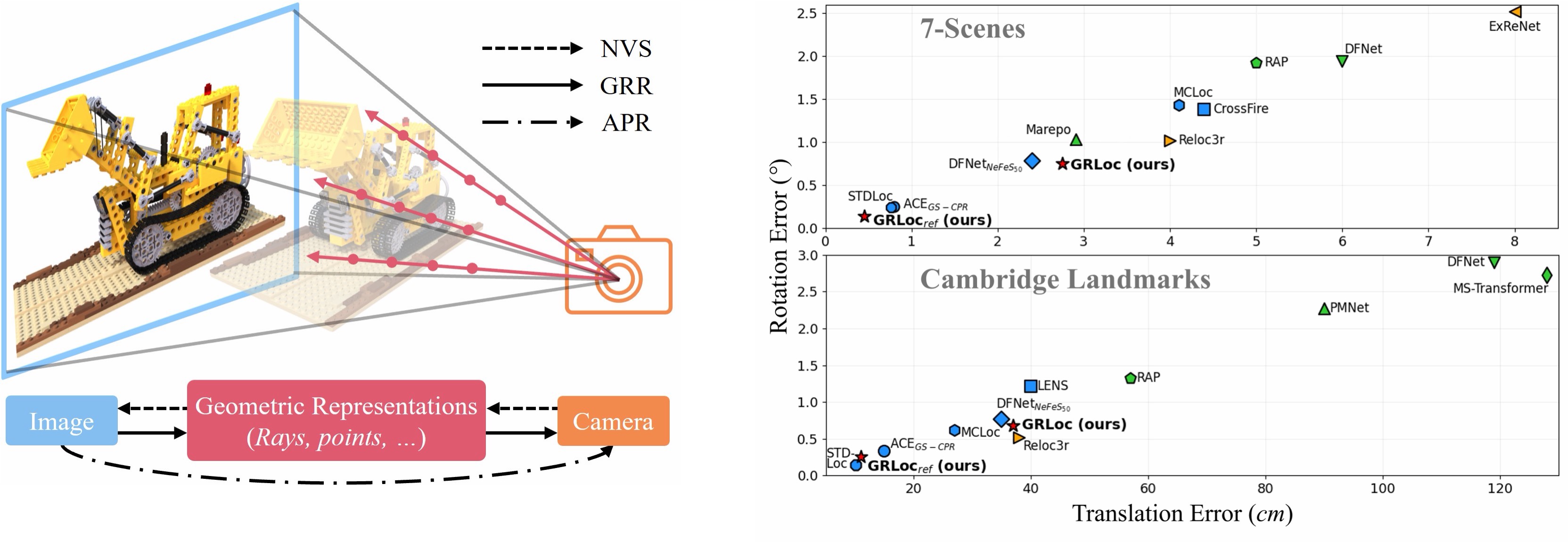}
    % \vspace{-6mm}
    \caption{Left: Our Geometric Representation Regression (GRR) paradigm is inspired by Novel View Synthesis (NVS). While NVS performs the forward rendering process, GRR learns the inverse. Right: Localization errors on the 7-Scenes~\cite{shotton2013scene} and Cambridge~\cite{kendall2015posenet} datasets. Our GRLoc outperforms prior \green{APR (green)} methods and demonstrates competitive performance against leading \orange{RPR (orange)} and \blue{PPR (blue)} approaches. Our refined model GRLoc\textsubscript{ref} achieves comparable results with SOTA \blue{PPR} methods.}
    % \vspace{-4mm}
    \label{fig:comparison}
\end{figure}

The paradigm of Absolute Pose Regression (APR), where a single neural network learns to predict a camera's 6-DoF pose directly from a query image, holds immense promise for simple and efficient visual localization systems~\cite{brahmbhatt2018geometry, kendall2015posenet, shavit2021learning}. This end-to-end approach contrasts with traditional methods that rely on complex, multi-stage pipelines of feature matching and geometric verification. However, the apparent simplicity of APR has been challenged by a critical finding: these networks often fail to generalize to novel viewpoints, instead functioning as sophisticated memory banks that retrieve the poses of the most visually similar training images~\cite{sattler2019understanding}. This tendency to ``memorize'' rather than ``understand'' suggests that the models are learning an opaque mapping from pixels to pose, bypassing the foundational geometric principles of camera projection.

This reveals a fundamental issue not in the network's capacity, but in how the task is formulated. Tasking a model with learning the direct, highly non-linear transformation from a 2D image to a 6-DoF pose vector is challenging because it encourages the network to discover statistical shortcuts. While synthetic data augmentation~\cite{moreau2022lens,lin2024learning,chen2022dfnet,li2024unleashing} can expand the ``memory bank'', it does not fundamentally change the black-box nature of the learning process.

The evolution of novel view synthesis (NVS) offers an instructive parallel. Recent methods like Neural Radiance Field (NeRF)~\cite{mildenhall2021nerf} and 3D Gaussian Splatting (3DGS)~\cite{kerbl20233d} decompose rendering into explicit geometric steps. Given camera parameters, these methods cast \textbf{rays} into the scene, sample \textbf{points} along rays or query Gaussians, accumulate appearance, and render the final image. This explicit, geometry-centric decomposition has proven far more effective at generating high-fidelity views. This success has inspired methods to invert the process, such as iNeRF~\cite{yen2021inerf}, which iteratively refines a pose via analysis-by-synthesis.

In this paper, we propose a paradigm shift that applies this principle in reverse for pose estimation. Just as NeRF and 3DGS decompose the forward rendering process, we decompose the inverse problem into a single-pass regression task, fundamentally different from iNeRF~\cite{yen2021inerf}: given an image, we first regress its underlying geometric representations, and then compute the camera pose from this structure. This idea is illustrated in Figure~\ref{fig:comparison}. Specifically, our network predicts two geometric representations for each image patch: a \textbf{raymap}'s directions to estimate camera rotation, and a \textbf{pointmap} to estimate camera translation.

We name this paradigm Geometric Representation Regression (GRR), and name our visual localization approach GRLoc. While our method's internal structure is different, it maintains the end-to-end simplicity of APR: it takes an image as input and produces a 6-DoF pose as output. The camera pose is then deterministically recovered from these geometric representations using a differentiable solver. Since the forward rendering in NeRF/3DGS is mathematically uninvertible in principle, this practical inversion requires learned perception to estimate geometric representations from image context. We leverage 3DGS to synthesize diverse training views, expanding scene coverage without requiring further real-world data collection.

A critical finding of this work is that decoupling the predictions for rotation and translation is key to improving performance. While balancing the optimization of both targets in a single branch is theoretically feasible, it is practically challenging. Our model design thus splits the network into a ray branch for rotation prediction and a point branch for translation prediction.

This hybrid framework separates learned perception (image to geometry) from analytical computation (geometry to pose), introducing a strong geometric prior and improving interpretability. Through domain-adversarial training to align synthetic and real data, our method achieves robust generalization. Our main contributions are: 
\begin{enumerate} 
\item We reformulate absolute pose estimation as an inverse rendering task, decomposing the problem into a learned perception step (image to geometry) and an analytical solving step (geometry to pose).

\item We introduce GRLoc that predicts 3D rays and points, and then converts them to a pose via a differentiable solver. GRLoc achieves state-of-the-art performance on the 7-Scenes and Cambridge Landmarks datasets.

\item We find that rotation and translation are competing objectives, and demonstrate that our decoupled geometric representations and network designs are critical for avoiding this conflict and achieving high accuracy.
\end{enumerate}

\section{Related Work}
\label{sec:relatedwork}

\subsection{Visual Localization Paradigms} 

Camera pose estimation is a long-standing problem in computer vision. A well-studied category of geometry-based methods establish correspondences between 2D keypoints in a query image and a sparse point cloud~\cite{liu2017efficient, dusmanu2019d2, lindenberger2023lightglue, sarlin2020superglue, taira2018inloc, giang2024learning, humenberger2020robust, sarlin2021back, sattler2011fast, sattler2016efficient, zeisl2015camera, sarlin2019coarse}, and use these 2D-3D matches to solve for the 6-DoF pose via PnP~\cite{gao2003complete} algorithms within a RANSAC~\cite{fischler1981random} loop. These methods achieve high accuracy due to their explicit geometric reasoning, but require storing and querying large 3D point clouds, raising concerns about storage and computation costs.

A more recent family of work, Scene Coordinate Regression (SCR)~\cite{brachmann2017dsac, brachmann2021visual, tang2021learning, brachmann2023accelerated, wang2024glace, jiang2025r}, replaces the reliance on local feature matching by training a neural network to directly predict dense, pixel-aligned 3D world coordinates for the query image. SCR methods maintain the geometric robustness of correspondence-based approaches while using compact implicit scene representations. Notable works include ACE~\cite{brachmann2023accelerated}, which introduced efficient scene-centric training, and GLACE~\cite{wang2024glace}, which improved scalability to large scenes. However, SCR methods still face challenges, including the reliance on a time-consuming geometric solver step, typically RANSAC-PnP~\cite{fischler1981random,gao2003complete}.

Another paradigm is Relative Pose Regression (RPR)~\cite{abouelnaga2021distillpose, arnold2022map, balntas2018relocnet, dong2025reloc3r, ding2019camnet, winkelbauer2021learning}, which regresses the relative pose between the query image and the most similar images in the scene database as references. RPR methods can leverage the rich information from reference images to improve pose accuracy, particularly in challenging scenarios. A main limitation is the image retrieval step needed for preparing the reference images, which can be time-consuming, especially in large scene databases, and the reliance on reference image quality.

In contrast, Absolute Pose Regression (APR)~\cite{brahmbhatt2018geometry, kendall2015posenet, shavit2021learning,blanton2022structure, chen2021direct, shavit2022camera, chen2022dfnet, lin2024learning, li2024unleashing, chen2024map} aims for a fully end-to-end solution, training a deep neural network to directly map an input image to a 6-DoF pose. The primary advantage of APR is its inference speed and simplicity, as it collapses a multi-stage pipeline into a single forward pass. However, this end-to-end solution has historically come at the cost of accuracy and interpretability, with APR models often achieving lower accuracy than geometry-based methods. The fundamental challenge lies in the compact nature of the pose representation: a 6-parameter vector is highly sensitive to noise and provides limited redundancy for optimization. This issue has been addressed from several perspectives, such as augmenting the model with larger-scale~\cite{moreau2022lens, chen2022dfnet, lin2024learning} or more diverse~\cite{li2024unleashing} training data.

Instead, we fundamentally reconsider the architecture of the regressor itself. We propose that the network's objective should not be to predict the final pose directly, but rather to infer intermediate geometric representations from which the pose can be recovered through well-established geometric algorithms. Our approach is inspired by the explicit reasoning of classical methods, but formulates the geometry prediction as a dense, learned task suitable for end-to-end training. Our method acts as an ``inverse camera,'' deconstructing an image into its constituent 3D components: a raymap and a pointmap. This disentangled `image $\rightarrow$ geometry $\rightarrow$ pose' framework contrasts with the `image $\rightarrow$ pose' paradigm of APR, and the `image $\rightarrow$ 3D coordinates $\rightarrow$ pose' pipeline of SCR, offering a new path toward building more robust and interpretable pose estimators.

Although the network's direct output differs from conventional APR, our GRR method remains fully end-to-end. The model's input (an image) and final output (a 6-DoF pose) are identical to APR, as the intermediate geometric representations are passed to a fully differentiable solver. Furthermore, the primary training objective aligns with APR: optimizing for the correctness of the final pose. Notably, while recent methods like Marepo~\cite{chen2024map} are recognized as APRs\cite{liu2024gs,sidorov2025gsplatloc} despite explicitly predicting scene coordinates (SCR style), GRLoc avoids dense coordinate maps by estimating camera-specific, patch-level geometry, making it conceptually closer to conventional APR. 

A recent work CRR~\cite{zhang2025semantic} also investigates ray regression. Our GRR differs in three key aspects: (1) We aim to "invert" rendering to learn generalizable 3D geometry, whereas CRR focuses on resolving the privacy and unbounded error issues of SCR. (2) CRR operates as a non-end-to-end SCR pipeline relying on RANSAC-PnP. (3) CRR utilizes Plücker coordinates, which entangle rotation and translation. Conversely, GRR explicitly decouples them following a design choice we found critical for optimal performance.

\subsection{Novel View Synthesis for Localization}

Recent advancements in novel view synthesis (NVS) have demonstrated impressive capabilities in reconstructing and rendering 3D scenes. Neural Radiance Fields (NeRF)~\cite{mildenhall2021nerf} pioneered the use of neural implicit representations to encode scene geometry and appearance in the weights of a multilayer perceptron, enabling photorealistic rendering of novel viewpoints through volumetric ray marching. Building on this foundation, 3D Gaussian Splatting (3DGS)~\cite{kerbl20233d} introduced an explicit representation using 3D Gaussians, achieving real-time rendering speeds while maintaining high visual fidelity through efficient rasterization. These NVS methods have opened new opportunities for enhancing visual localization in multiple ways.

First, NVS provides a powerful data augmentation mechanism for expanding limited training datasets. APR methods have successfully leveraged NeRF~\cite{moreau2022lens,lin2024learning} and 3DGS~\cite{chen2022dfnet,li2024unleashing} to synthesize large quantities of novel views with accurate ground-truth poses, addressing the data scarcity problem that has long hindered learning-based localization. This synthetic-to-real transfer has proven particularly valuable for training robust pose regressors, though it introduces the challenge of bridging the domain gap between rendered and real images.

Another family of work, Post-Pose Refinement (PPR) methods~\cite{germain2022feature,liu2024hr,trivigno2024unreasonable}, aims to iteratively refine an initial coarse pose estimate. Many recent PPR methods also leverage NVS~\cite{chen2024neural,moreau2023crossfire,zhao2024pnerfloc,zhou2024nerfect,liu2024gs,huang2025sparse,sidorov2025gsplatloc,pietrantoni2025gaussian}. These methods pass an initial pose prediction to an NVS model, render a synthetic view from that estimated viewpoint, and refine the predicted pose by minimizing the photometric or feature-based discrepancy between the rendered view and the query image. This ``render and compare'' strategy effectively converts the pose estimation problem into an optimization process, achieving impressive refinement accuracy when starting from reasonable initial estimates.

Our method also utilizes 3DGS, drawing inspiration from both approaches. First, similar to APR methods, we leverage 3DGS to generate synthetic training views that significantly expand our training data beyond the limited real-world images available for a given scene, allowing our model to learn more robust geometric representations by observing the scene from diverse viewpoints. Second, we incorporate 3DGS-based post-refinement following the settings of RAP~\cite{li2024unleashing} as an optional step to further improve the pose accuracy. However, our primary contribution lies in the initial pose estimation via GRR, not the refinement process itself. While we demonstrate that our method is compatible with post-refinement, we treat it as an optional step to fairly evaluate our core approach.

\section{Method}

We formalize the absolute pose estimation problem as follows. Given a single RGB image $\mathbf{I} \in \mathbb{R}^{H \times W \times 3}$, our goal is to estimate the camera pose represented by a camera-to-world transformation matrix $\mathbf{T} \in \mathrm{SE}(3)$, which includes a rotation matrix $\mathbf{R} \in \mathrm{SO}(3)$ and a translation vector $\mathbf{t} \in \mathbb{R}^3$ representing the camera center in world coordinates.

% \begin{equation}
% \mathbf{T} = \begin{bmatrix} \mathbf{R} & \mathbf{t} \\ \mathbf{0}^\top & 1 \end{bmatrix} \in \mathbb{R}^{4 \times 4}
% \end{equation}

\begin{wrapfigure}[15]{r}{0.3\linewidth}
  % \vspace{-1mm}
	\includegraphics[width=1.0\linewidth]{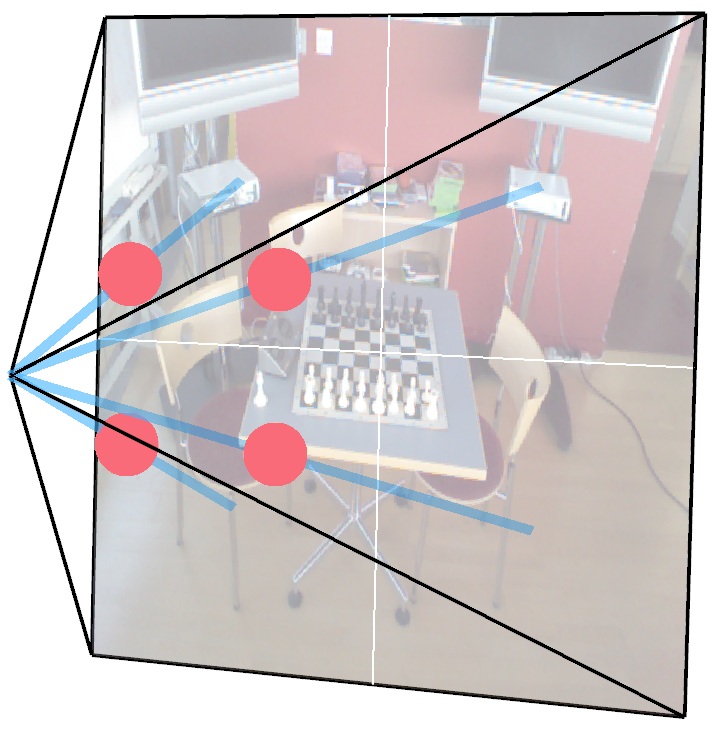}
	\caption{Geometric representations for an example $4 \times 4$ grid: (1) a raymap of view directions and (2) a pointmap of 3D points.}
	\label{fig:modeling}
\end{wrapfigure}

Rather than directly regressing the pose parameters, we reformulate this problem as estimating intermediate geometric representations that encode the 3D structure of the camera's view. Specifically, we decompose the image into a grid of $N \times N$ patches. 

While the forward rendering process of NVS (casting rays to form pixels) is well-defined, its inverse is not. Our goal of ``inverting'' this process, which is to estimate 3D representations from a 2D image, is a learned perception task. It requires considering both local patch information and global spatial relationships, a task that aligns well with Vision Transformer~\cite{dosovitskiy2020image} architectures. We predict two types of \textbf{geometric representations} for each patch $i$, as illustrated in Figure~\ref{fig:modeling}:

\begin{itemize}
    \item \textbf{Ray direction field} $\hat{\mathcal{D}} = \{ \hat{\mathbf{d}}_i \in \mathbb{S}^2 \}_{i=1}^{N \times N}$: A set of predicted unit vectors in world coordinates representing the viewing direction corresponding to patch $i$. Each patch in the grid acts as a 'super-pixel', and its predicted ray $\hat{\mathbf{d}}_i$ represents the average direction of all pixel-rays within that region. For simplicity, we refer to these as ``rays'', though they technically represent direction vectors.
    
    \item \textbf{3D pointmap} $\hat{\mathcal{P}} = \{ \hat{\mathbf{p}}_i \in \mathbb{R}^3 \}_{i=1}^{N \times N}$: A set of predicted points in world coordinates. Each point $\hat{\mathbf{p}}_i$ is modeled to lie on the corresponding ray $\hat{\mathbf{d}}_i$ and is constrained to be at a unit distance from the camera's origin ($\|\hat{\mathbf{p}}_i - \mathbf{t}\| = 1$), which removes scale ambiguity. This constraint defines the pointmap in a camera-relative canonical space (like directional vectors) and allows recovering translation without regressing unbounded world coordinates.
\end{itemize}

Given these predicted representations $\hat{\mathcal{D}}$ and $\hat{\mathcal{P}}$, and their known canonical (in camera space) counterparts $\mathcal{D}^{\text{cam}}$ and $\mathcal{P}^{\text{cam}}$, our goal is to recover the camera pose. As our ablation studies (Section~\ref{sec:ablation}) suggest, rotation and translation can be competing objectives when entangled. We therefore solve for them using two specialized, differentiable alignment problems. At a high level, this involves solving for the transformation $\mathbf{T}$ that minimizes the joint error:

\begin{equation}
\mathbf{T} = \underset{\mathbf{R}, \mathbf{t}}{\arg\min} \sum_{i=1}^{N^2} \| \mathbf{R} \mathbf{p}_i^{\text{cam}} - \mathbf{d}_i \|^2 + \| \mathbf{R} \mathbf{p}_i^{\text{cam}} + \mathbf{t} - \mathbf{p}_i \|^2
\end{equation}
where the first term formulates a rotation-only alignment problem for the rays, and the second term formulates a rigid 3D registration problem for the points.

We discretize the problem by predicting one representation per patch in an $N \times N$ grid. While a denser grid (i.e., 1 pixel per patch) could theoretically improve the solver's robustness by providing more constraints, this benefit is conditional on the network's ability to predict these dense representations accurately. Therefore, selecting the patch grid density requires a trade-off between geometric detail coverage and per-patch prediction quality.

\begin{figure*}[t]
    \centering
    \includegraphics[width=1.0\textwidth]{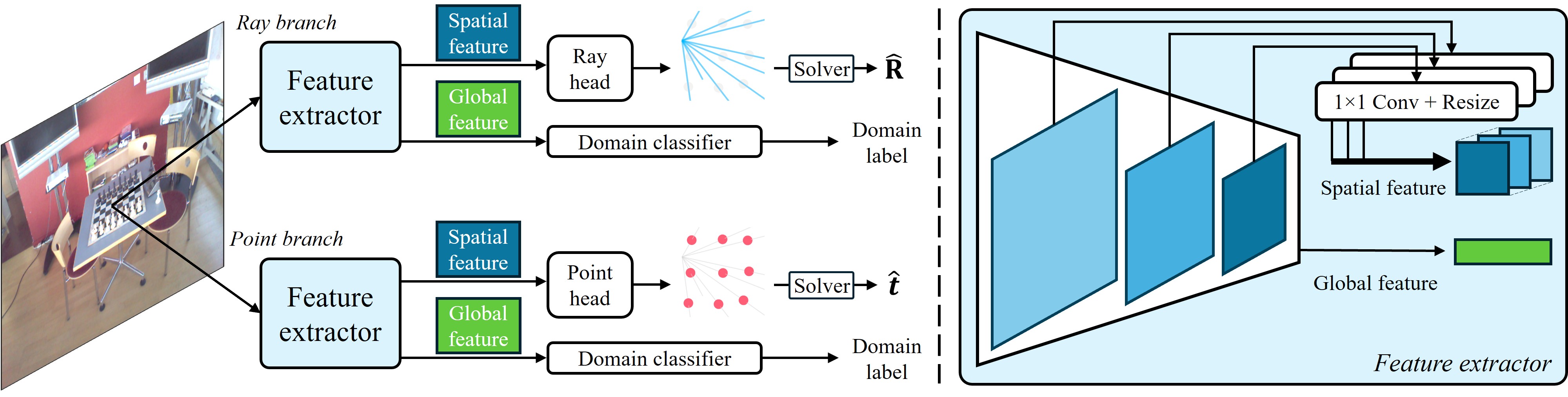}
    \caption{An overview of our proposed architecture. Left: the query image is fed into a decoupled dual-branch network. Each branch's feature extractor produces spatial features for a prediction head and a global feature for a domain classifier. The \textbf{ray head} and \textbf{point head} predict their respective geometric representations. The final pose components are then analytically recovered from these representations using differentiable deterministic solvers. Right: The feature extractor fuses multi-level features from the backbone via $1 \times 1$ convolutions, resizing, and concatenation to produce the spatial feature, while the backbone's original final output is used as the global feature.}
    \label{fig:overview}
\end{figure*}

An overview of our architecture is shown in~\autoref{fig:overview}. The input image is processed by a dual-branch network, which is intentionally decoupled to support this specialized alignment. The ray branch predicts per-patch ray directions, while the point branch predicts per-patch 3D points. These predictions are fed into their respective deterministic, differentiable solvers, allowing us to leverage their complementary strengths in the final loss: we use the rotation computed from the ray alignment and the translation computed from the point alignment.

\subsection{Geometric Representation Prediction}
Our model's architecture is intentionally designed as a decoupled dual-branch network to predict the two distinct geometric representations from a single RGB image. This design prevents the competing objectives of rotation and translation from interfering with each other at the feature level.

\paragraph{Feature Extractors.}
Each of the \textit{ray branch} and the \textit{point branch} has its own feature extractor built with a transformer-based backbone and a fusion neck. The extractor's job is to produce two outputs: a single global feature $\mathbf{f}_{global}$ for domain alignment, and a spatially-aware multi-level feature map $\mathbf{F}_{spatial}$ for the prediction heads.

A feature extractor uses a transformer-based backbone, which is well-suited for our per-patch prediction task of geometric representation prediction. To capture rich local information, each branch's multi-level feature maps are processed by a dedicated fusion neck, as shown in \autoref{fig:overview} (Right). This neck unifies their channel dimensions via $1\times1$ convolutions, resizes them to a target spatial resolution of $N \times N$, and concatenates them. This produces a branch-specific, spatially-aware feature tensor $\mathbf{F}_{spatial}$, where each spatial location corresponds to an image patch.

\paragraph{Prediction Heads.}
Two lightweight MLP heads, \textit{ray head} and \textit{point head}, regress the geometric representations from their respective branch's spatial features $\mathbf{F}_{spatial}$, and output the dense ray direction field $\hat{\mathcal{D}}$ and 3D pointmap $\hat{\mathcal{P}}$.

\subsection{Deterministic Pose Solver}
A key aspect of our method is that the final camera pose is not directly predicted. Instead, we estimate it deterministically from the predicted geometric representations using non-learnable but fully differentiable solvers. This step offloads the complex, non-linear pose estimation from the network and allows for end-to-end training. As our design is decoupled, we compute two separate pose estimates to provide specialized supervision for rotation and translation.

Both of our alignment tasks (the rotation-only Procrustes problem for rays and the rigid point set registration for points) are solved using the Kabsch algorithm~\cite{kabsch1976solution}, which provides a closed-form solution for finding the optimal rigid transformation between two sets of corresponding vectors or points. At its core, the algorithm relies on a Singular Value Decomposition (SVD) of a covariance matrix. This makes the solver highly efficient and, crucially for our end-to-end framework, fully differentiable.

\paragraph{Rotation from Rays.} The ray branch is specialized for rotation. We solve a rotation-only Procrustes problem to find the optimal rotation $\hat{\mathbf{R}} \in \mathrm{SO}(3)$ that best aligns the predicted world-space ray directions $\hat{\mathcal{D}}$ with the canonical ray directions $\mathcal{D}^{\text{cam}}$. This approach is critical because the raymap is a pure representation of rotation, free from the competing objective of translation.

\paragraph{Translation from Points.} We find the full transformation $[\hat{\mathbf{R}}_{\text{point}} | \hat{\mathbf{t}}_{\text{point}}]$ that best aligns the predicted 3D point cloud $\hat{\mathcal{P}}$ with the canonical point cloud $\mathcal{P}^{\text{cam}}$. As suggested by our ablation studies in Section~\ref{sec:ablation}, this pointmap is an entangled representation: while it can solve for a complete pose, its rotation estimate is less accurate than the one from the pure ray branch. We therefore enforce our decoupling strategy by using only the translation component $\hat{\mathbf{t}}_{\text{point}}$ as the whole model's prediction $\hat{\mathbf{t}}$ from this solved transformation.

\subsection{Data Augmentation with Novel View Synthesis}
Similar to prior APR methods~\cite{moreau2022lens,lin2024learning,chen2022dfnet,li2024unleashing}, we leverage NVS to augment our limited training datasets by rendering novel views to expand scene coverage. We specifically use 3D Gaussian Splatting (3DGS)~\cite{kerbl20233d} for this task due to its fast and efficient rendering. Our process is performed offline: we first train a 3DGS model on the real training images, and then use it to render a large, fixed set of synthetic views. Our final training data thus consists of the original real images and the much larger synthetic set. In each training epoch, we iterate through the synthetic set once while repeatedly cycling through the smaller real set.

Training on synthetic data introduces a domain gap between the rendered and real images. To bridge this gap, we employ domain-adversarial adaptation~\cite{ganin2016domain,tzeng2017adversarial}. As shown in \autoref{fig:overview}, a Gradient Reversal Layer (GRL) is applied to the global feature $\mathbf{f}_{global}$ from each branch. This feature is then fed to a domain classifier that is trained to discriminate between the real and synthetic domains. The GRL forces the backbones to learn domain-invariant features that can ``fool'' this classifier. This allows our model to benefit from training on the massive set of augmented views while maintaining its ability to generalize to real-world images.

\subsection{Training and Objective Functions}
\label{sec:loss}
Our model is trained end-to-end with a composite loss function that provides comprehensive supervision on both the final pose estimates and the intermediate geometric representations. Following a warmup phase, we switch the norm $p$ of loss from 1 to 2 to refine the predictions more precisely.

\paragraph{Pose Supervision Loss.} We supervise both recovered poses against the ground truth pose $\mathbf{T}_{gt} = [\mathbf{R}_{gt} | \mathbf{t}_{gt}]$. The loss combines rotation and translation errors:

\begin{equation}
    \mathcal{L}_{pose} = w^{\text{r}}_{pose}d_g(\hat{\mathbf{R}}, \mathbf{R}_{gt})^p + w^{\text{p}}_{pose}\|\hat{\mathbf{t}} - \mathbf{t}_{gt}\|_p
\end{equation}
where $d_g$ is the geodesic distance between two rotations.

\paragraph{Geometry Loss.} To ensure our intermediate representations are geometrically meaningful, we directly supervise them against their ground-truth counterparts. The loss is a sum of two terms: (1) a ray term that encourages alignment between the predicted rays $\hat{\mathcal{D}}$ and ground-truth rays $\mathcal{D}_{gt}$ using cosine similarity, and (2) a point term that penalizes the distance between the predicted points $\hat{\mathcal{P}}$ and the ground-truth points $\mathcal{P}_{gt}$:

\begin{equation}
    \mathcal{L}_{geo} = w^{\text{r}}_{geo}\left(1 - \cos(\hat{\mathcal{D}}, \mathcal{D}_{gt})\right) + w^{\text{p}}_{geo}\|\hat{\mathcal{P}} - \mathcal{P}_{gt}\|_p
\end{equation}

\paragraph{Geometric Regularization Loss.} We introduce two regularization terms to enforce geometric smoothness across the predicted fields, in order to: (1) preserve local angular relationships and (2) preserves pairwise distances for local rigidity:

\begin{align}
\mathcal{L}_{\text{reg}} = \frac{1}{|\mathcal{N}|}\sum_{(i,j)\in\mathcal{N}} & w^{\text{r}}_{reg}\Big| (\hat{\mathbf{d}}_i \cdot \hat{\mathbf{d}}_j) - (\mathbf{d}_i^{\text{cam}} \cdot \mathbf{d}_j^{\text{cam}}) \Big|_p \nonumber \\
+ &w^{\text{p}}_{reg}\Big| \|\hat{\mathbf{p}}_i - \hat{\mathbf{p}}_j\| - \|\mathbf{p}_i^{\text{gt}} - \mathbf{p}_j^{\text{gt}}\| \Big|_p
\end{align}
where $\mathcal{N}$ denotes the set of spatial patch neighbors.

\paragraph{Domain Adaptation Loss.}
We follow the domain-adversarial training via gradient reversal~\cite{ganin2016domain} on both ray and point cloud branches. The loss for a given batch is the sum of the Binary Cross-Entropy (BCE) losses:

\begin{equation}
\mathcal{L}_{\text{domain}} = \sum_{b \in \{\text{ray}, \text{point}\}} \mathcal{L}_{\text{BCE}}(D^b(\text{GRL}(\mathbf{f}^b_{global})), y)
\end{equation}
where $D^b$ is the domain classifier for a branch, and $y$ is the domain label (e.g., $0$ for synthetic, $1$ for real).

\paragraph{Total Loss.} The final training objective is a weighted sum of these components, calculated for both a batch of synthetic data ($\mathcal{L}_{syn}$) and a batch of real data ($\mathcal{L}_{real}$), along with the domain adaptation loss:
\begin{align}
    \mathcal{L}_{total} = & w_{syn}\mathcal{L}^{syn} + w_{real}\mathcal{L}^{real} \nonumber \\
    & + w_{domain}(\mathcal{L}^{syn}_{domain} + \mathcal{L}^{real}_{domain})
\end{align}
where $\mathcal{L}^{syn/real} = \mathcal{L}_{pose} + \mathcal{L}_{geo} + \mathcal{L}_{reg}$.

\section{Experiments}

We discuss our experimental setup, implementation details, and results in this section. As discussed, our GRR paradigm aligns with APR in its end-to-end design and primary training objective. The key difference is our intermediate regression of geometric representations. Therefore, we categorize our method with APR for our main comparisons.

\begin{figure*}[t]
    \centering
    \includegraphics[width=1.0\textwidth]{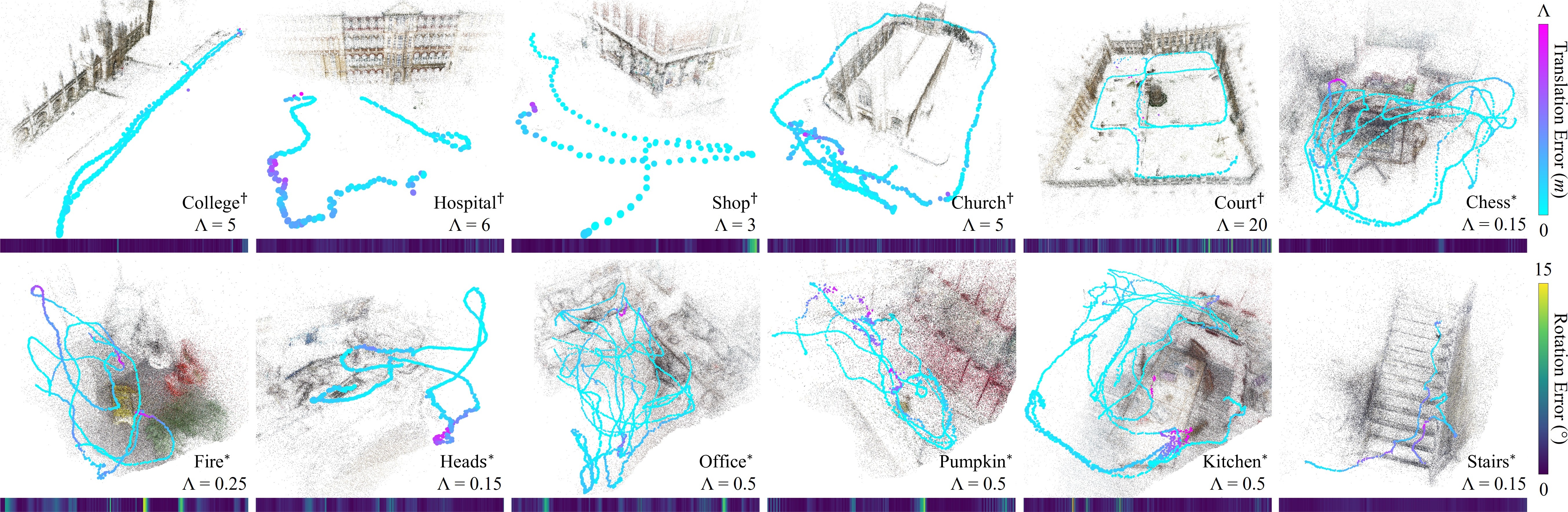}
    \vspace{-6mm}
    \caption{Visualization of GRLoc's predictions. $\dagger$: Cambridge Landmarks~\cite{kendall2015posenet}; $*$: 7-Scenes~\cite{shotton2013scene}. $\Lambda$ indicates scene-specific upper-limit of translation error. Sequential error bars show rotation errors.}
    \vspace{-2mm}
    \label{fig:visualization}
\end{figure*}

\begin{figure*}[t]
    \centering
    \includegraphics[width=1.0\textwidth]{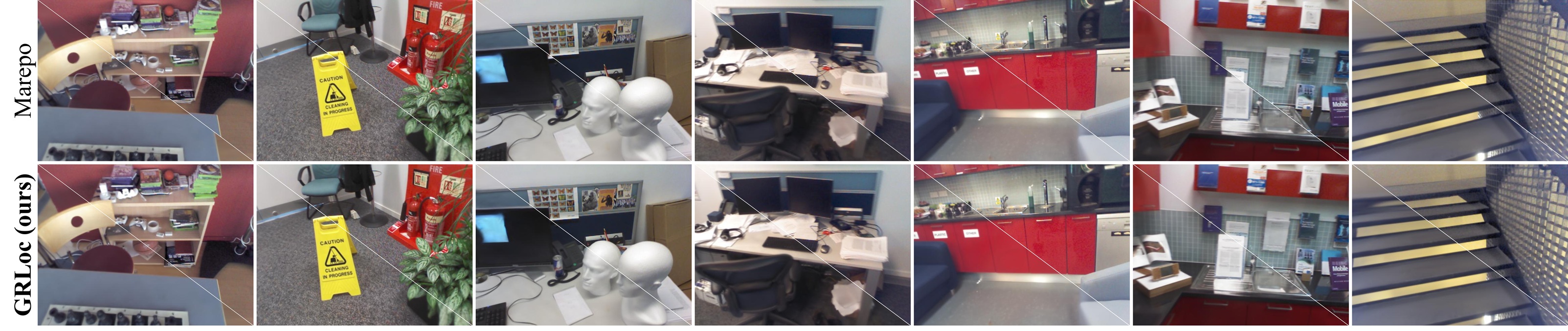}
    \vspace{-6mm}
    \caption{Visual comparisons on 7-Scenes dataset~\cite{shotton2013scene}: top-right corners show ground truth, and bottom-left corners show estimated views.}
    \vspace{-2mm}
    \label{fig:renders}
\end{figure*}

\subsection{Experimental Setup}

\paragraph{Datasets.} In our experiments, we use two public datasets for evaluation: (1) the Cambridge Landmarks dataset~\cite{kendall2015posenet}, which consists of large-scale outdoor urban scenes ranging approximately $875\sim9200m^2$, and (2) the 7-Scenes dataset~\cite{shotton2013scene}, which consists of small-scale indoor environments with spatial extents ranging $1\sim18m^2$. For fair comparisons with APR methods, we follow the setup of recent works~\cite{chen2022dfnet,lin2024learning,li2024unleashing} to sample subsets of the training data.

% $50\%$ for Cambridge, and either $10\%$ or $20\%$ for 7-Scenes (depending on whether a training sequence exceeds $2000$ frames).

\paragraph{Baselines.} Our primary comparison is with state-of-the-art Absolute Pose Regression (APR) methods, including PMNet~\cite{lin2024learning}, DFNet~\cite{chen2022dfnet}, RAP~\cite{li2024unleashing}, and Marepo~\cite{chen2024map}. For a comprehensive evaluation, we also report results against leading RPR, SCR, and PPR methods. 

While our main focus is single-frame localization, we also test an optionally refined model $\text{GRLoc}_{\text{ref}}$ to ensure a fair comparison against high-accuracy SCR and PPR methods, which include refinement or non-differentiable solver steps. We adopt a recent pose-refinement strategy~\cite{liu2024gs,li2024unleashing} and follow the exact refinement setup of RAP~\cite{li2024unleashing}.

\paragraph{Implementation Details.} All of our experiments are conducted on a machine with 32 AMD EPYC 9654 CPU cores and a single NVIDIA H200 GPU.

Our feature extractor's backbone is the Swin Transformer V2 large~\cite{liu2022swin}, pretrained on ImageNet~\cite{deng2009imagenet}, to leverage its multi-level feature maps. We empirically set the patch grid shape to $16 \times 16$, which provides the best balance across different grid resolutions for the discussed density-performance trade-off.

For loss weights, our preliminary tests suggest using fixed weights can cause specific terms to dominate depending on diverse scene scales. We thus employ a scale-adaptive weighting strategy to ensure balanced gradient flow, initializing weights to normalize the initial magnitude to 1.0.

For 3DGS-based data augmentation, we use the gsplat library~\cite{ye2025gsplat} and follow the 3DGS-MCMC~\cite{kheradmand20243d} method. We generate the synthetic dataset with sampled novel camera poses by perturbing the original training views.

\subsection{Evaluations on Benchmarks}
Figures~\ref{fig:visualization} and ~\ref{fig:renders} show qualitative results on benchmarks. Additional visualizations are provided in supplement. 

On the room-scale indoor scenes of the 7-Scenes dataset (using SfM pseudo ground truth~\cite{brachmann2021limits}), our approach achieves state-of-the-art performance among APR/GRR methods, as shown in Table~\ref{tab:7scenes_results}. Notably, our $\text{GRLoc}$ outperforms Marepo~\cite{chen2024map} on average and in 5 scenes, though Marepo is trained with additional priors from the Map-Free dataset~\cite{arnold2022map}.

On the large-scale outdoor scenes of the Cambridge Landmarks dataset, $\text{GRLoc}$ significantly outperforms previous APR methods, with an average reduction of over $33\%$ in translation error and $46\%$ in rotation error, as shown in Table~\ref{tab:cambridge_results}. Our method also demonstrates strong performance on the challenging \textit{Court} scene, a large-scale environment that is often omitted by prior APR work.

\begin{table*}[t]
\tiny
% \fontsize{5}{6}\selectfont
\caption{Median translation (cm) and rotation (°) errors on the 7-Scenes dataset~\cite{shotton2013scene}. Bold: best results; Underline: second-best.}
\vspace{-2mm}
\centering
\resizebox{\textwidth}{!}{%
\begin{tabular}{ll|ccccccc|c}
\hline
\textbf{Category} & \textbf{Methods} & \textbf{Chess} & \textbf{Fire} & \textbf{Heads} & \textbf{Office} & \textbf{Pumpkin} & \textbf{Kitchen} & \textbf{Stairs} & \textbf{Average} \\
\hline
& PoseNet~\cite{kendall2015posenet} & 10/4.02 & 27/10.00 & 18/13.00 & 17/5.97 & 19/4.67 & 22/5.91 & 35/10.50 & 21/7.74 \\
& MapNet~\cite{brahmbhatt2018geometry} & 13/4.97 & 33/9.97 & 19/16.7 & 25/9.08 & 28/7.83 & 32/9.62 & 43/11.8 & 28/10.00 \\
& PAE~\cite{shavit2022camera} & 13/6.61 & 24/12.00 & 14/13.00 & 19/8.58 & 17/7.28 & 18/8.89 & 30/10.30 & 19/9.52 \\
\textbf{APR}  & MS-Transformer~\cite{shavit2021learning} & 11/6.38 & 23/11.5 & 13/13.0 & 18/8.14 & 17/8.42 & 16/8.92 & 29/10.30 & 18/9.51 \\
& DFNet~\cite{chen2022dfnet} & 3/1.12 & 6/2.30 & 4/2.29 & 6/1.54 & 7/1.92 & 7/1.74 & 12/2.63 & 6/1.93 \\
& RAP~\cite{li2024unleashing} & \underline{2/0.85} & 6/2.84 & 4/4.52 & \underline{4}/1.57 & 3/1.10 & 5/1.10 & 10/\underline{1.30} & 5/1.90 \\
& Marepo~\cite{chen2024map} & 1.9/0.83 & \textbf{2.3}/\textbf{0.92} & \underline{2.1/1.24} & \textbf{2.9}/\textbf{0.93} & \underline{2.5/0.88} & \underline{2.9/0.98} & \underline{5.9}/1.48 & \underline{2.9/1.04} \\
\textbf{(GRR)} & \textbf{GRLoc (ours)} & \textbf{0.97/0.31} & \underline{2.93}/\underline{0.95} & \textbf{1.40/0.90} & 5.08/\underline{1.05} & \textbf{2.28/0.53} & \textbf{3.13/0.69} & \textbf{3.42/0.82} & \textbf{2.75/0.75} \\
\hline
\textbf{RPR} & ExReNet~\cite{winkelbauer2021learning} & 5/1.63 & 7/2.54 & 3/2.71 & 6/1.75 & 7/2.04 & 7/2.10 & 19/4.87 & 8/2.52 \\
& Reloc3r~\cite{dong2025reloc3r} & 3/0.88 & 3/0.81 & 1/0.95 & 4/0.88 & 6/1.10 & 4/1.26 & 7/1.26 & 4/1.02 \\
\hline
& DSAC~\cite{brachmann2021visual} & 0.5/0.17 & 0.8/0.28 & 0.5/0.33 & 1.2/0.34 & 1.2/0.28 & 0.7/0.21 & 2.7/0.78 & 1.1/0.34 \\
\textbf{SCR} & ACE~\cite{brachmann2023accelerated} & 0.5/0.18 & 0.8/0.33 & 0.6/0.34 & 1.0/0.29 & 1.0/0.22 & 0.8/0.20 & 2.9/0.81 & 1.1/0.34 \\
& GLACE~\cite{wang2024glace} & 0.6/0.18 & 0.9/0.34 & 0.5/0.33 & 1.1/0.29 & 0.9/0.23 & 0.8/0.20 & 3.2/0.93 & 1.2/0.36 \\
\hline
& FQN-MN~\cite{germain2022feature} & 4.1/1.31 & 10.5/2.97 & 7.63/5.86 & 3.6/2.36 & 4.6/1.76 & 16.1/4.42 & 139.5/34.67 & 28/7.3 \\
& CrossFire~\cite{moreau2023crossfire} & 1/0.4 & 3/2.3 & 5/1.9 & 5/1.6 & 3/0.8 & 2/0.8 & 12/1.9 & 4.4/1.38 \\
& MCLoc~\cite{trivigno2024unreasonable} & 2/0.8 & 3/1.4 & 2/1.45 & 4/1.3 & 5/1.6 & 6/1.6 & 6/2.0 & 4.1/1.43 \\
& HR-APR~\cite{liu2024hr} & 2/0.55 & 2/0.75 & 3/1.3 & 2/0.64 & 2/0.62 & 2/0.67 & 5/1.30 & 2.4/0.85 \\
\textbf{PPR} & DFNet + NeFeS\textsubscript{50}~\cite{chen2024neural} & 2/0.57 & 2/0.74 & 2/1.28 & 2/0.56 & 2/0.55 & 2/0.57 & 5/1.28 & 2.4/0.79 \\

& NeRFMatch~\cite{zhou2024nerfect} & 0.95/0.30 & 1.11/0.41 & 1.34/0.92 & 3.09/0.87 & 2.21/0.60 & 1.03/0.28 & 9.26/1.74 & 2.71/0.73 \\
% & DFNet+GS-CPR~\cite{liu2024gs} & 0.7/0.20 & 0.6/0.36 & 0.9/0.32 & 1.2/0.32 & 1.3/0.31 & 0.9/0.25 & 2.2/0.61 & 1.1/0.34 \\
& ACE + GS-CPR~\cite{liu2024gs} & 0.5/0.15 & 0.4/0.28 & 0.6/0.25 & 0.9/0.26 & 1.0/0.23 & 0.7/0.17 & 1.4/0.42 & 0.8/0.25 \\
& STDLoc~\cite{huang2025sparse} & 0.46/0.15 & 0.57/0.24 & 0.45/\underline{0.26} & 0.86/0.24 & 0.93/0.21 & 0.63/0.19 & 1.42/0.41 & 0.76/0.24 \\
& RAP\textsubscript{ref}~\cite{li2024unleashing} & \underline{0.33/0.11} & \underline{0.51/0.21} & \underline{0.39}/0.27 & \textbf{0.57}/\underline{0.16} & \underline{0.81/0.20} & \underline{0.45/0.12} & \underline{1.11/0.32} & \underline{0.60/0.20} \\
& \textbf{GRLoc\textsubscript{ref} (ours)} & \textbf{0.28/0.10} & \textbf{0.32/0.13} & \textbf{0.22/0.14} & \underline{0.65}/\textbf{0.15} & \textbf{0.70/0.15} & \textbf{0.35/0.10} & \textbf{0.61/0.19} & \textbf{0.45/0.14} \\
\hline
\end{tabular}
}
\vspace{-3mm}
\label{tab:7scenes_results}
\end{table*}

\begin{table}[t]
\fontsize{6}{8}\selectfont
% \scriptsize
\caption{Median translation (cm) and rotation (°) errors on the Cambridge~\cite{kendall2015posenet}. Bold: best results; Underline: second-best.}
\vspace{-2mm}
\centering
\setlength{\tabcolsep}{2pt}
\begin{tabular}{ll|cccc|c|c}
\hline
& \textbf{Methods} & \textbf{College} & \textbf{Hospital} & \textbf{Shop} & \textbf{Church} & \textbf{Average} & \textbf{Court} \\
\hline
& PoseNet~\cite{kendall2015posenet} & 166/4.86 & 262/4.90 & 141/7.18 & 245/7.95 & 204/6.23 & 683/3.50 \\
& MapNet~\cite{brahmbhatt2018geometry} & 107/1.89 & 194/3.91 & 149/4.22 & 200/4.53 & 163/3.64 & N/A \\
& PAE~\cite{shavit2022camera} & 90/1.49 & 207/2.58 & 99/3.88 & 164/4.16 & 140/3.03 & N/A \\
\textbf{APR} & MS-Trans.~\cite{shavit2021learning} & 83/1.47 & 181/2.39 & 86/3.07 & 162/3.99 & 128/2.73 & N/A \\
& DFNet~\cite{chen2022dfnet} & 73/2.37 & 200/2.98 & 67/2.21 & 137/4.03 & 119/2.90 & 217/4.11 \\
& PMNet~\cite{lin2024learning} & 68/1.97 & 103/1.31 & 58/2.10 & 133/3.73 & 90/2.27 & N/A \\
& RAP~\cite{li2024unleashing} & \underline{52/0.90} & \underline{87/1.21} & \underline{33/1.48} & \underline{53/1.52} & \underline{56/1.28} & \underline{115/1.68} \\
\textbf{(GRR)} & \textbf{GRLoc (ours)} & \textbf{34/0.42} & \textbf{54/0.88} & \textbf{15/0.48} & \textbf{46/0.93} & \textbf{37/0.68} & \textbf{88/0.53} \\
\hline
\textbf{RPR} & ExReNet~\cite{winkelbauer2021learning} & 233/2.48 & 354/3.49 & 72/2.41 & 230/3.72 & 222/3.03 & 979/4.46 \\
& Reloc3r~\cite{dong2025reloc3r} & 42/0.36 & 62/0.55 & 13/0.58 & 34/0.58 & 38/0.52 & 122/0.73 \\
\hline
& DSAC~\cite{brachmann2021visual} & 18/0.3 & 21/0.4 & 5/0.3 & 15/0.6 & 15/0.4 & 34/0.2 \\
\textbf{SCR} & ACE~\cite{brachmann2023accelerated} & 28/0.4 & 31/0.6 & 5/0.3 & 18/0.6 & 21/0.5 & 43/0.2 \\
& GLACE~\cite{wang2024glace} & 19/0.3 & 17/0.4 & 4/0.2 & 9/0.3 & 12/0.3 & 19/0.1 \\
\hline
& LENS~\cite{moreau2022lens} & 34/0.54 & 45/0.96 & 28/1.66 & 54/1.66 & 40/1.21 & N/A \\
& FQN-MN~\cite{germain2022feature} & 28/0.38 & 54/0.82 & 13/0.63 & 58/2.00 & 38/0.96 & 4253/9.16 \\
& CrossFire~\cite{moreau2023crossfire} & 47/0.7 & 43/0.7 & 20/1.2 & 39/1.4 & 37/1.0 & N/A \\
& NeFeS\textsubscript{50}~\cite{chen2024neural} & 37/0.54 & 52/0.88 & 15/0.53 & 37/1.14 & 35/0.77 & N/A \\
& HR-APR~\cite{liu2024hr} & 36/0.58 & 53/0.89 & 13/0.51 & 38/1.16 & 35/0.78 & N/A \\
\textbf{PPR} & MCLoc~\cite{trivigno2024unreasonable} & 31/0.42 & 39/0.73 & 12/0.45 & 26/0.88 & 27/0.62 & N/A \\
& NeRFMatch~\cite{zhou2024nerfect} & 12.5/0.23 & 20.9/0.38 & 8.4/0.40 & 10.9/0.35 & 14.5/0.29 & 19.6/0.09 \\
% & DFNet\textsubscript{GS-CPR}~\cite{liu2024gs} & 23/0.32 & 42/0.74 & 10/0.36 & 27/0.62 & 26/0.51 & N/A \\
& ACE\textsubscript{GS-CPR}~\cite{liu2024gs} & 20/0.29 & 21/0.40 & 5/0.24 & 13/0.40 & 15/0.33 & N/A \\
& DFNet\textsubscript{ref}~\cite{li2024unleashing} & 16/0.24 & 21/0.41 & 8/0.42 & 10/0.26 & 14/0.33 & 25/0.13 \\
& STDLoc~\cite{huang2025sparse} & \textbf{15.0/0.17} & \textbf{11.9/0.21} & \textbf{3.0/0.13} & \textbf{4.7/0.14} & \textbf{10.1/0.14} & \textbf{15.7/0.06} \\
& RAP\textsubscript{ref}~\cite{li2024unleashing} & \textbf{15}/0.23 & 18/0.38 & \underline{5}/0.23 & 9/\underline{0.23} & 12/0.27 & 22/0.15 \\
& \textbf{GRLoc\textsubscript{ref} (ours)} & 16/\underline{0.21} & \underline{16/0.35} & \underline{5/0.20} & \underline{8/0.23} & \underline{11/0.25} & \underline{21/0.11} \\
\hline
\end{tabular}
\vspace{-4mm}
\label{tab:cambridge_results}
\end{table}

The results also demonstrate that our base model is already competitive with or superior to several PPR methods on both datasets. When applying the optional refinement step, $\text{GRLoc}_{\text{ref}}$ achieves results comparable to $\text{RAP}_{\text{ref}}$~\cite{li2024unleashing}. This is expected, as we adopt the same refinement module. We observe that while our unrefined $\text{GRLoc}$ outperforms the unrefined $\text{RAP}$, both methods provide a sufficiently accurate initial pose for the refinement module to converge to a similar, high-quality result. During these experiments, we also noted that the refinement performance is highly dependent on the quality of the depth-enabled 3DGS model: the fidelity of its rendered RGB influences the 2D feature matching, while the accuracy of its rendered depth directly impacts the final 2D-to-3D correspondence estimation.

Interestingly, while $\text{GRLoc}_{\text{ref}}$ outperforms STDLoc~\cite{huang2025sparse} on 7-Scenes, it performs slightly worse on Cambridge. We attribute this to the challenges of large-scale outdoor environments, which are not only more difficult for the pose estimator itself, but also for training the high-fidelity NVS models. As discussed, the refinement quality is highly dependent on the NVS model's fidelity, and it is more difficult to achieve this in large outdoor scenes.

\subsection{Inference Latency and Real-Time Feasibility}

To evaluate the computational efficiency of our approach, we measure the inference latency in an online setting with a batch size of 1. Our $\text{GRLoc}$ achieves an average inference time of $36.78$ ms per frame. While slower than the highly lightweight RAP~\cite{li2024unleashing} ($12.84$ ms), it remains significantly faster than Marepo~\cite{chen2024map} ($68.50$ ms). The increased latency compared to RAP is primarily attributed to the high-capacity Swin Transformer backbone~\cite{liu2022swin}, which is computationally heavier but critical for driving our accuracy gains. Despite this heavy backbone, $\text{GRLoc}$ maintains a competitive speed that satisfies the requirements of most real-time applications, such as augmented reality and autonomous navigation.

\subsection{Ablation Study}
\label{sec:ablation}
We conduct our ablation studies on the \textit{Shop} scene from the Cambridge Landmarks dataset to evaluate different model designs. The results are shown in Table~\ref{tab:ablation}. We define the following experimental conditions:

\begin{itemize}
    \item \textbf{ViT} replaces the Swin Transformer backbone with a vanilla ViT L-16~\cite{dosovitskiy2020image}.
    \item \textbf{W/o DA} removes the domain adaptation components during training.
    \item \textbf{APR} uses the same architecture as the full model, but is trained to directly regress the 6-DoF pose from the global features following standard APR.
    \item \textbf{GR-only} trains the network using only the geometric representation (raymap and pointmap) supervision, without utilizing the final 6-DoF pose as a direct loss signal. The final pose is purely recovered by the solver during inference.
    \item \textbf{W/o point} and \textbf{W/o ray} keep only the ray or point branch, respectively. \textbf{W/o point} adds a MLP head to regress the 3D translation, as the ray branch cannot solve for it.
    \item \textbf{Share} uses a shared feature extractor for both heads.
\end{itemize}

As shown in Table~\ref{tab:ablation}, we first validate our architectural choices. A Swin Transformer backbone with multi-level feature fusion successfully outperforms the \textbf{ViT} backbone. Furthermore, the significant performance drop in the \textbf{W/o DA} highlights the necessity of domain adaptation for robust feature extraction. 

To validate our core formulation, we compare different supervision paradigms: \textbf{APR} (pure pose supervision), \textbf{GR-only} (pure geometric supervision), and our full model. The results indicate that pairing explicitly learned geometric representations with final pose supervision yields the optimal performance.

Beyond the supervision paradigm, finding a robust representation for both rotation and translation is critical. The comparison between \textbf{W/o point}, \textbf{W/o ray}, \textbf{Share}, and our full \textbf{GRLoc} model traces the development of our approach:

\begin{enumerate}
    \item We first considered a ray-only model (\textbf{W/o point}), which provides a pure representation for rotation. However, this model lacks a geometric basis for translation, forcing us to add a direct regression head that performs poorly.
    \item We then tested a point-only model (\textbf{W/o ray}), with pointmap as an entangled representation. This model achieved better translation, but its rotation error was worse than the ray-only variant. This confirms our hypothesis that an entangled representation is suboptimal for recovering rotation.
    \item Next, we combined both representations in a \textbf{Share} model using a single backbone, outperformed the single-branch variants but was still suboptimal. We found that the two competing objectives were ``pulling against'' each other, forcing the shared features to be a poor compromise for both tasks.
\end{enumerate}

\begin{table}[t]
\scriptsize
\caption{Ablation studies on the \textit{Shop} scene. Median translation (cm) and rotation (\textdegree) errors are reported.}
\vspace{-2mm}
\centering
\begin{tabular}{ll|ll|ll|ll}
\hline
\textbf{ViT}       & 29/0.98 & \textbf{W/o DA}  & 33/1.25 & \textbf{APR}   & 39/1.67 & \textbf{GR-only}      & 20/3.96 \\ \hline
\textbf{W/o point} & 41/0.66 & \textbf{W/o ray} & \textbf{15}/0.86 & \textbf{Share} & 17/0.77 & \textbf{GRLoc (ours)} & \textbf{15/0.48} \\ \hline
\end{tabular}
\vspace{-2mm}
\label{tab:ablation}
\end{table}

This led to our final GRLoc model that fully decouples the two branches. This design allows each branch to extract features specialized for its unique objective, leading to the best performance. These observations validate our claim: it is critical to decouple rotation and translation, occurring at both the representation level (by using a pure geometric representation like ray directions for rotation) and the architectural level (by using separate network branches to prevent competing objectives from degrading the features). 

We believe the conflict stems from the contradictory nature of required visual features: rotation relies on translation-invariant global cues (e.g., vanishing points), while translation requires parallax-sensitive local cues (e.g., relative scale). In entangled backbones, the network must learn features that are simultaneously invariant to and sensitive to position, causing the model to ``cheat'' by using translation shifts to compensate for rotational drift.

\section{Conclusion}

In this paper, we present a reformulation of absolute pose estimation as Geometric Representation Regression (GRR), a new paradigm that regresses explicit geometric components and then analytically computes the 6-DoF pose using a differentiable solver. We demonstrate that decoupling the rotation and translation predictions is critical for resolving their competing optimization objectives and achieving high accuracy. Our GRLoc achieves state-of-the-art performance, validating that this hybrid, geometry-centric approach provides a more robust, interpretable, and generalizable path forward for visual localization.

\bibliographystyle{splncs04}
\bibliography{main}
\end{document}